\documentclass{article}

\usepackage{PRIMEarxiv}

\usepackage[utf8]{inputenc} 
\usepackage[T1]{fontenc}    
\usepackage{hyperref}       
\usepackage{url}            
\usepackage{booktabs}       
\usepackage{amsfonts}       
\usepackage{nicefrac}       
\usepackage{microtype}      
\usepackage{lipsum}
\usepackage{fancyhdr}       
\usepackage{graphicx}       
\usepackage{subfigure}
\usepackage{algorithm}
\usepackage{algorithmic}
\usepackage{amsmath,amssymb}
\usepackage{bm}
\usepackage{multirow}
\usepackage{makecell}
\usepackage{booktabs}
\usepackage{color}
\usepackage{colortbl}
\graphicspath{{media/}}     

\title{A Deep Model for Partial Multi-Label Image Classification with Curriculum Based Disambiguation
}

\author{
  Feng Sun, Ming-Kun Xie and Sheng-Jun Huang \\
  College of Computer Science and Technology, Nanjing University of Aeronautics and Astronautics \\
  MIIT Key Laboratory of Pattern Analysis and Machine Intelligence, Nanjing, 211106, China \\
  \texttt{\{sunfeng, mkxie, huangsj\}@nuaa.edu.cn} \\
}

\begin{document}

\def \x {\bm{x}}
	\def \y {\bm{y}}
	\def \p {\bm{p}}
	\def \w {\bm{w}}
	\def \btheta {\bm{\theta}}
	\def \W {\bm{W}}
	\def \bomega {\bm{\omega}}
	\definecolor{mygray}{gray}{.9}
	
\maketitle

\begin{abstract}
In this paper, we study the partial multi-label (PML) image classification problem, where each image is annotated with a candidate label set consists of multiple relevant labels and other noisy labels. Existing PML methods typically design a disambiguation strategy to filter out noisy labels by utilizing prior knowledge with extra assumptions, which unfortunately is unavailable in many real tasks. Furthermore, because the objective function for disambiguation is usually elaborately designed on the whole training set, it can be hardly optimized in a deep model with SGD on mini-batches. In this paper, for the first time we propose a deep model for PML to enhance the representation and discrimination ability. On one hand, we propose a novel curriculum based disambiguation strategy to progressively identify ground-truth labels by incorporating the varied difficulties of different classes. On the other hand, a consistency regularization is introduced for model retraining to balance fitting identified easy labels and exploiting potential relevant labels. Extensive experimental results on the commonly used benchmark datasets show the proposed method significantly outperforms the SOTA methods.
\end{abstract}


\section{Introduction}
\label{sec:intro}

Multi-label image classification is a commonly used framework for object recognition with complex semantics, where each image is associated with multiple class labels. For example, a single X-ray scan can detect multiple suspected pathologies such
as tuberculosis and pneumonia \cite{ge2018chest}. The goal of multi-label image classification is to train a multi-label classifier that can accurately recognize a set of objects presented in an unseen image.

\begin{figure}[!t]
	\centering
	\includegraphics[width=0.9\textwidth]{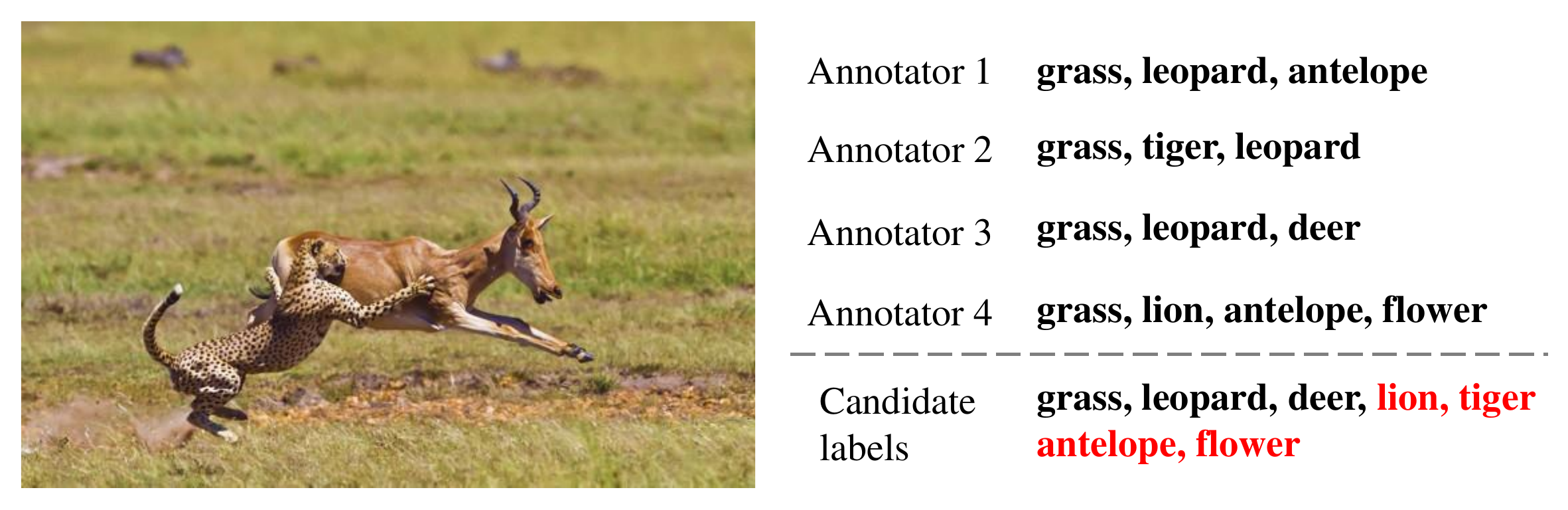}
	\caption{An example of partial multi-label image classification. The image is partial-labeled by annotators with different level of expertise on the crowdsourcing platform. In the candidate label set, grass, leopard and deer are ground-truth labels while lion, tiger, antelope and flower are noisy labels.}
	\label{fig:example}
\end{figure}

In conventional multi-label image classification, a common assumption is that each training image has been precisely annotated with all of its relevant labels. Unfortunately, in real-world scenarios, it is difficult and costly to obtain a large number of precise annotations. An alternative solution is to assign a candidate label set to each image by non-expertise labelers. Besides the relevant labels, the candidate label set also contains some irrelevant labels (also called noisy labels). For example (as shown in Figure \ref{fig:example}), in the crowdsourcing scenario, the image may be annotated with a candidate label set, which is the union set of annotations from multiple labelers with different level of expertise. While the labeling cost is significantly reduced, the learning problem becomes more challenging, since the ground-truth labels are confused with noisy ones. To solve this problem, a learning framework called partial multi-label learning (PML) \cite{xie2018partial} (or partial multi-label image classification (PMLC) in our paper) is proposed and has been widely studied \cite{sun2019partial,zhang2020pml,wang2019discriminative,xu2021progressive,lyu2020partial}. 

A simple method to solve the PMLC problem is to treat all candidate labels as relevant ones and convert the original problem into a standard multi-label image classification problem. Then the converted problems can be solved by employing the off-the-shelf multi-label image classification approaches, such as ML-GCN~\cite{chen2019multi}, ASL~\cite{ridnik2021asymmetric}, Query2Label~\cite{liu2021query2label}. Unfortunately, the performance of these methods may be significantly degraded, since they are misled by the noisy labels in the candidate set.

To mitigate this issue, a large number of methods have been developed to handle partial labels by adopting the disambiguation strategy, which aims to identify the relevant labels in the candidate set. The main idea is to perform disambiguation for candidate labels based on either prior knowledge, such as structure information of data \cite{sun2019partial,zhang2020pml}, or auxiliary information, such as additional validation data \cite{xie2021partial2}.

Although  disambiguation strategies have shown improvement of practical performance for solving PMLC tasks, there still exist two main challenges for learning a multi-label image classifier on large-scale PMLC datasets. On one hand, existing methods highly depend on the prior knowledge or auxiliary information, which may be unavailable in many real-world scenarios and thus leads the model to obtain an unfavorable performance. On the other hand, the elaborate objective function for disambiguation is usually defined on the whole dataset, and can be hardly optimized with a deep model, further leading to lack of strong representation and discrimination ability. This also makes it impossible to train deep neural networks by fine-tuning the powerful pre-trained models, which has been examined in our experiments to be important for obtaining a desirable performance.

In this paper, we propose to learn a deep neural network for solving PMLC tasks by performing curriculum based disambiguation with consistency regularization (CDCR). Specifically, we progressively identify easy labels in the candidate set with high confidence by incorporating with the varied difficulties of different classes. Furthermore, the consistency regularization is introduced to retrain the model by conducting stochastic augmentation for training images, aiming to boost the disambiguation performance. Extensive experimental results show that it significantly outperforms traditional disambiguation strategies and obtains state-of-the-art performance on the commonly-studied PMLC benchmark datasets. Specifically, CDCR obtains 89.0\% mAP on VOC with 5.3 average candidate labels (contains 1.6 average true labels) compare to the previous state-of-the-art of 84.8\% as well as the fully supervised performance of 90.4\%.

The rest of this paper is organized as follows: Section \ref{sec:related work} reviews related works. Section \ref{sec:approach} introduces our proposed CDCR approach, followed by experimental results in Section \ref{sec:experiment}. Finally, we conclude in Section \ref{sec:conclusion} with discussions on future work.

\section{Related Work}
\label{sec:related work}

Partial multi-label image classification (PMLC) is particularly relevant to two popular learning frameworks, i.e., multi-label image classification \cite{wang2016cnn} and partial multi label learning \cite{xie2018partial}.

In multi-label image classification (MLC), each image is associated with multiple class labels simultaneously. Existing MLC methods can be roughly classified into three categories. The first kind of methods attempt to capture the correlations among labels, which have been regarded as a fundamental element to obtain a desirable model performance. Among them, ML-GCN \cite{chen2019multi} and its variants are representative methods which utilize the GCN to capture the pairwise correlation between labels. The second kind of methods focus on designing specific loss functions to tackle the positive-negative imbalance issue in multi-label classification tasks \cite{ridnik2021asymmetric}. Given the fact that an image often contains multiple objects, the third kind of methods aim to locate areas of interest related to semantic labels by using attention technique \cite{liu2021query2label}.

To solve PML problems, a large number of methods have been developed based on the disambiguation strategy, i.e., identify true labels in the candidate set. Some of them perform disambiguation by estimating a confidence for each candidate label, and then employ an alternative optimization scheme to update confidences and train a multi-label classifier based on the estimated confidences \cite{xie2018partial,sun2019partial}. Other methods  design a two-stage strategy to primarily transform the PML task into a supervised multi-label learning problem by constructing a label representation for each partial-label example \cite{xu2020partial,zhang2020pml}. Then, a multi-label classifier can be easily trained on the transformed data. However, when dealing with partial multi-label images, these methods often fail to obtain desirable performances due to the lack of strong representation and discrimination abilities, since they often train a linear model based on hand-craft features. 

The proposed CDCR method is also relevant to curriculum learning. The basic idea of curriculum learning is to learn examples in a meaningful order, which involves harder example gradually. Early methods focus on designing easy-to-hard curriculums based on prior knowledge \cite{bengio2009curriculum}. 
Unfortunately, in many real-world scenarios, it is difficult and costly to obtain the prior information. Instead of pre-defined curriculums designed by human experts, the self-paced model is proposed to automatically identify easy examples based on their losses \cite{kumar2010self}. Recently, meta learning is utilized to learn a data-driven curriculum dynamically with a StudentNet \cite{jiang2018mentornet}. There exist a few of studies that deal with partial labels by borrowing the idea of curriculum learning. For example, in \cite{lyu2020partial2}, authors employ 
curriculum learning techniques to solve partial label learning (PLL) problems, which can be regarded as a special case of PML, where only one label is valid in the candidate set. Besides the studied setting, there still exist several significant differences compared to our work. Firstly, unlike our method, which designs a curriculum based disambiguation strategy for candidate labels, their work only considers curriculum by learning examples from easy to hard. Secondly, their work is designed specifically for linear model and thus cannot directly applied to solve PMLC problems. Besides, a recent work employs curriculum learning to handle the problem of multi-label learning with missing labels (MLML), where only a subset of true labels is partially known for training the model \cite{durand2019learning}. However, this method never consider performing model re-training based on the identified possible true labels by using the consistency regularization, which has been validated as a fundamental element to achieve desirable performance of PMLC in Section \ref{sec:consistency} and experiments. Additional, our method further improve the performance of curriculum based disambiguation by considering difficult of different classes.

\section{The Proposed Approach}
\label{sec:approach}

For each partial-labeled training image, we denote $\x_i$ as $i$-th training image and $\y_i$ as it corresponding label vector with $K$ possible class labels. We further use $\tilde{\y}_i$ to denote the candidate label vector for $\x_i$. In our setting, $\tilde{y}_{ij}=1$ yields the $j$-th label is a candidate label for image $\x_i$ while $\tilde{y}_{ij}=0$, otherwise. In the following content, the index $i$ will be omitted when it is clear by the context. Let $p(\y|\x)$ be the predicted probability distribution of true label vector $\y$ and $p(y_j|\x)$ be the predicted probability of $j$-th label for input $\x$. Furthermore, we denote $\btheta$ as the parameters of a deep neural network.

\begin{figure*}[!t]
	\centering
	\includegraphics[width=0.9\textwidth]{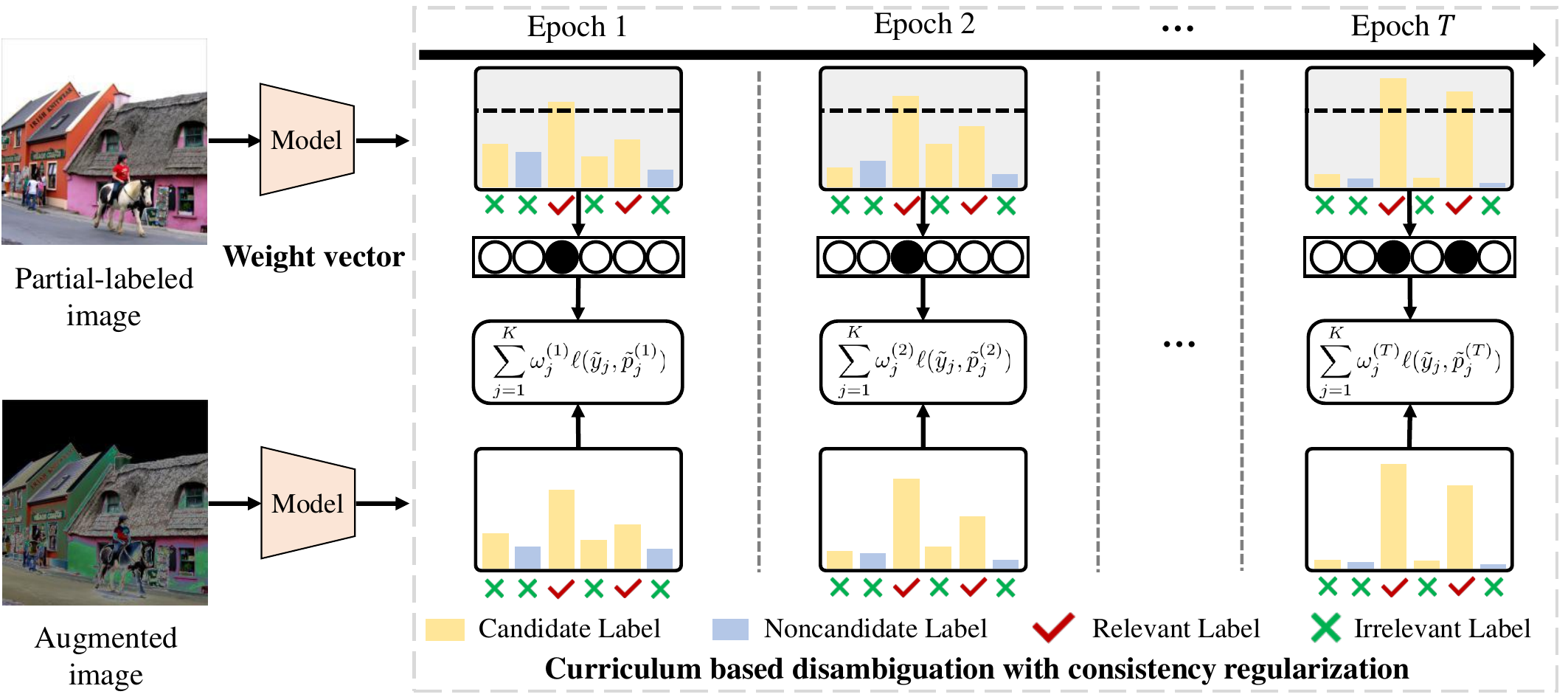}
	\caption{The illustration of our CDCR framework. CDCR strategy progressively identifies true labels and sets their weights as 1. The consistency regularization is used to retrain the model by conducting a stochastic augmentation for each training image.}
	\label{fig:framework}
\end{figure*}

Figure \ref{fig:framework} illustrates the disambiguation process of the proposed CDCR framework. For each partial-labeled image, CDCR utilizes a curriculum based disambiguation strategy to progressively identify true labels from easy to hard. Specifically, at early epochs, CDCR primarily identifies the easy candidate label (i.e., the third label) with a larger predicted probability than a threshold (marked by the black dashed line) and assigns its weight as 1 (the weights for other labels as 0). Then, CDCR retrains the model with the weighted BCE loss by performing consistency regularization  between the original partial-labeled image and its augmented version. As the disambiguation proceeds, harder true labels (i.e., the fifth label) can be identified, which significantly boosts the model performance. In the following content, we will introduce each component of the CDCR framework in detail.

\subsection{Loss Function}
In multi-label classification, the most commonly used loss is binary cross-entropy (BCE) loss, which decomposes the original task into multiple independent binary classification problems. By using $p_j$ to denote $p(y_j|\x)$, given candidate label vector $\tilde{\y}$, the multi-label BCE loss can be defined as:
\begin{equation}\label{equ: BCE_loss}
	L_{\text{BCE}} = -\sum_{j=1}^K[ \tilde{y}_j\log\left(p_j\right)
	+ (1- \tilde{y}_j)\log\left(1-p_j\right)].
\end{equation}

Unfortunately, the standard multi-label BCE loss usually introduces noisy labels, i.e, false positive labels, since it simply treats all candidate as relevant ones. These noisy labels often mislead the model and make it fail to obtain a desired performance. To mitigate this issue, for the candidate labels of training image $\x_i$, we introduce a binary weight vector $\bomega_i\in\{0, 1\}^q$, where the weight $\omega_{ij}$ indicates whether the $j$-th label is a relevant label for image $\x_i$. For non-candidate label $j$, since it is certainly irrelevant to image $\x_i$, it always satisfies $\omega_{ij}=0$. Accordingly, the weighted BCE loss can be formulated as follows:
\begin{equation}\label{eq:weight_loss}
	L= \frac{1}{n}\sum_{i=1}^n\sum_{j=1}^K\omega_{ij}\ell(\tilde{y}_{ij},p(y_j|\x_i)),
\end{equation}
where $\ell(y,p)=-[y\log(p)+(1-y)\log(1-p)]$ is the BCE loss. In the ideal case, if we have $w_{ij}=1$ when $\tilde{y}_j$ is a relevant label for image $\x_i$, then the true labels are identified, which indicates we have achieved disambiguation for candidate labels. 

\subsection{Curriculum based Disambiguation}
\label{sec:cd}
However, in many real-world scenarios, it is impossible to obtain the true weights $\bomega$ without any prior knowledge. In the following content, we will show how to estimate the true weights for the candidate labels. 

Specifically, we propose to perform disambiguation for the candidate labels with a curriculum based strategy, which progressively identifies easy candidate labels and treats them as true ones. For the notation simplicity, we denote $\Omega=\{\bomega_i\}_{i=1}^n$ as the set of weights for all training images. Then, the objective function for PMLC with curriculum based disambiguation (CD) can be formulated as follows:
\begin{equation}\label{eq:DC}
	L_{\text{CD}} = \frac{1}{n}\sum_{i=1}^n\sum_{j=1}^K\omega_{ij}\ell(\tilde{y}_{ij},p(y_j|\x_i)) +\Gamma(\Omega;\lambda)
\end{equation}
where the function $\Gamma(\Omega;\lambda)$ represents a curriculum, which is parameterized by $\lambda$. In order to obtain the optimal parameters $\btheta^*$, inspired of \cite{kumar2010self}, we employ an alternative optimization strategy, where each of $\btheta$ and $\Omega$ is alternatively minimized with the other being fixed. In the initial stage of training, it requires to warm up the model for a few of epochs by using partial labels with the standard BCE loss. Then, the disambiguation is performed for candidate labels by progressively identifying easy labels in the candidate set and then treats them as true labels for retraining the model.

Following the curriculum proposed in \cite{kumar2010self}, we define the function $\Gamma(\Omega;\lambda)$ in Eq.\eqref{eq:DC} as follows:
\begin{equation}\label{eq:SP}
	\Gamma(\Omega) = -\lambda\sum_{i=1}^N\sum_{j=1}^K \omega_{ij}
\end{equation}
When $\btheta$ is fixed, the optimal $\Omega$ can be obtained by Alternative Convex Search (ACS), which is an iterative method for biconvex optimization \cite{jiang2015self,gorski2007biconvex}. Specifically, the solution of $\Omega$ can be obtained as follows: $\omega_{ij} = 1$ if $\ell(\tilde{y}_{ij},p(y_j|\x_i)) < \lambda$, where $\lambda$ is a threshold for controlling the curriculum, while $\omega_{ij} = 0$, otherwise. Since for a label $\tilde{y}_{ij}$, a smaller loss yields a larger probability of $y_j$ to be true label for $x_i$, it is equivalent to progressively identify true labels based on their probabilities for the convenience of choosing the threshold. Therefore, the optimal $\Omega$ can be obtained as follows:
\begin{equation}\label{eq:curriculum}
	\omega^*_{ij} = {\mathbb I}(p(y_j|\x_i)\ge\alpha), \forall i\in[n], j\in\{j'|\tilde{y}_{ij'}=1\},
\end{equation}
 where ${\mathbb I}(\cdot)$ is the indicator function which outputs 1 if the condition holds while outputs 0, otherwise, and $\alpha$ is a threshold for controlling the disambiguation process.
 
 The intuition behind the CD strategy can be summarized as the following two folds. On one hand, recent studies \cite{arpit2017closer,han2018co} on the memorization effects of deep neural networks for learning with noisy labels disclose that the model would first memorize clean labels, i.e, output high probabilities for them. This means by adopting a suitable (large enough) value as the threshold, it is likely to identify true labels from the candidate set, which leads the model to be free of over-fitting to noisy labels. On the other hand, researches on curriculum learning \cite{bengio2009curriculum} show that training examples in a meaningful order, i.e., from easy to hard, can boost the model performance and robustness. Following the similar idea, primarily learning easy labels is beneficial for achieving disambiguation for hard labels in the candidate set.
 
 ~\\
\noindent\textbf{Difficulties of different classes.} In this section, we propose to perform disambiguation by incorporating difficulties of different classes. In general, the difficulties of different semantic classes are varied, which are mainly for the following two aspects. On one hand, the difficulty can be regarded as an intrinsic attribute for a specific class, which often depends on whether the semantic information for this class is easy to be distinguished. On the other hand, as mentioned in recent studies \cite{wu2020distribution}, in multi-label image classification, model training often suffers the class-imbalance issue, which makes the model over-fit to the head classes (which can be regarded as easy ones) with a large number of relevant labels while under-fit to the tail classes (which can be regarded as difficult ones) with a small number of relevant labels. As mentioned before, we perform disambiguation based on the predicted probability for each candidate label. Unfortunately, model often predict high probabilities for labels with respect to easy classes while output small probabilities for labels with respect to difficult classes, which makes the proposed CD strategy suffers a bias towards easy classes, i.e, identify much more labels from easy classes and eventually hardly identify labels from difficult classes. The phenomenon would significantly degrade the model performance due to the insufficient supervised information for difficult classes.

\begin{figure}[!tb]
	\centering
	
	\subfigure[Precision of identified labels]{
		\label{fig:precision}
		\centering
		\includegraphics[width=0.4\linewidth]{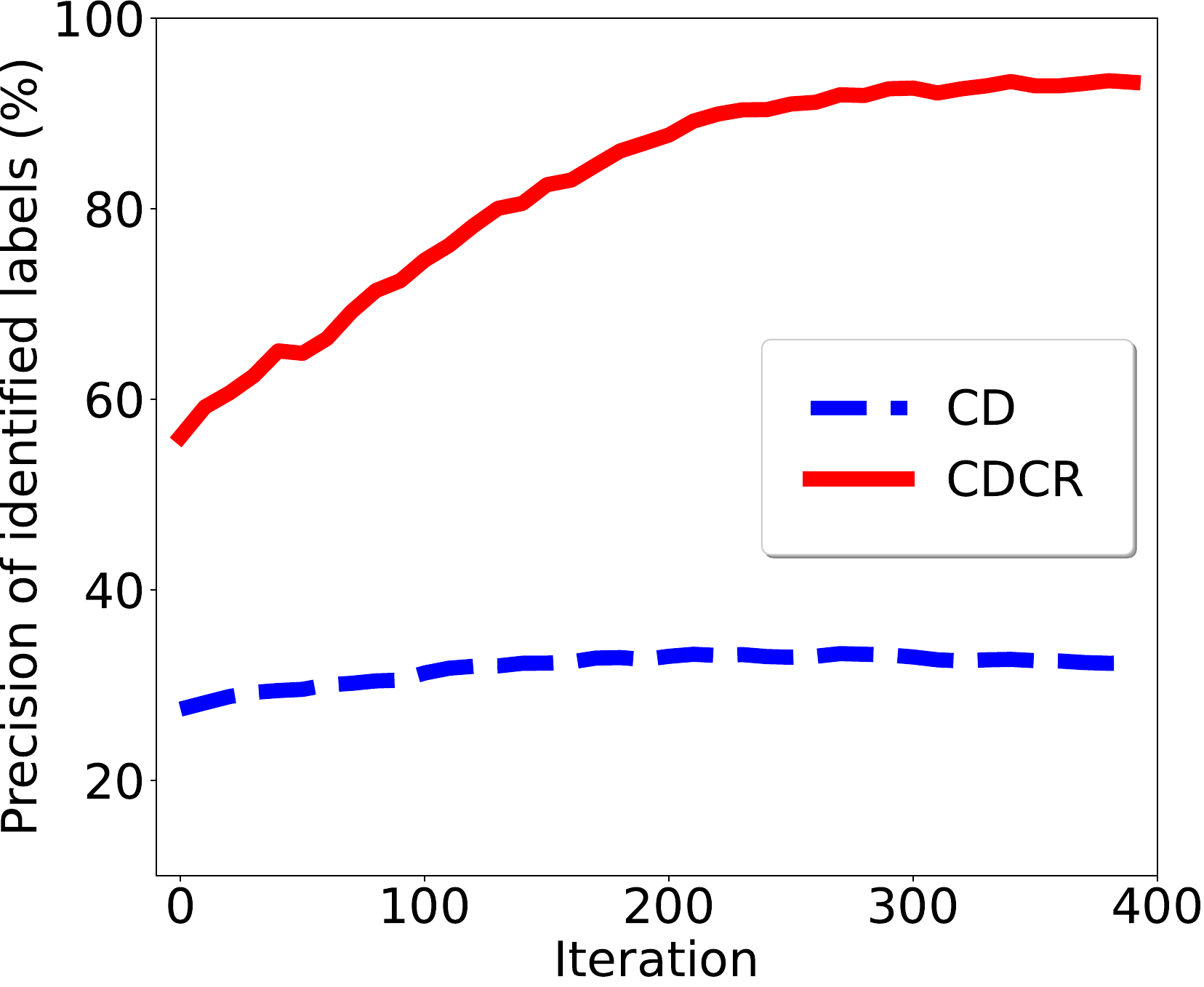}
	} 
	\subfigure[Number of repeated noisy labels]{
		\label{fig:repeat}
		\centering
		\includegraphics[width=0.4\linewidth]{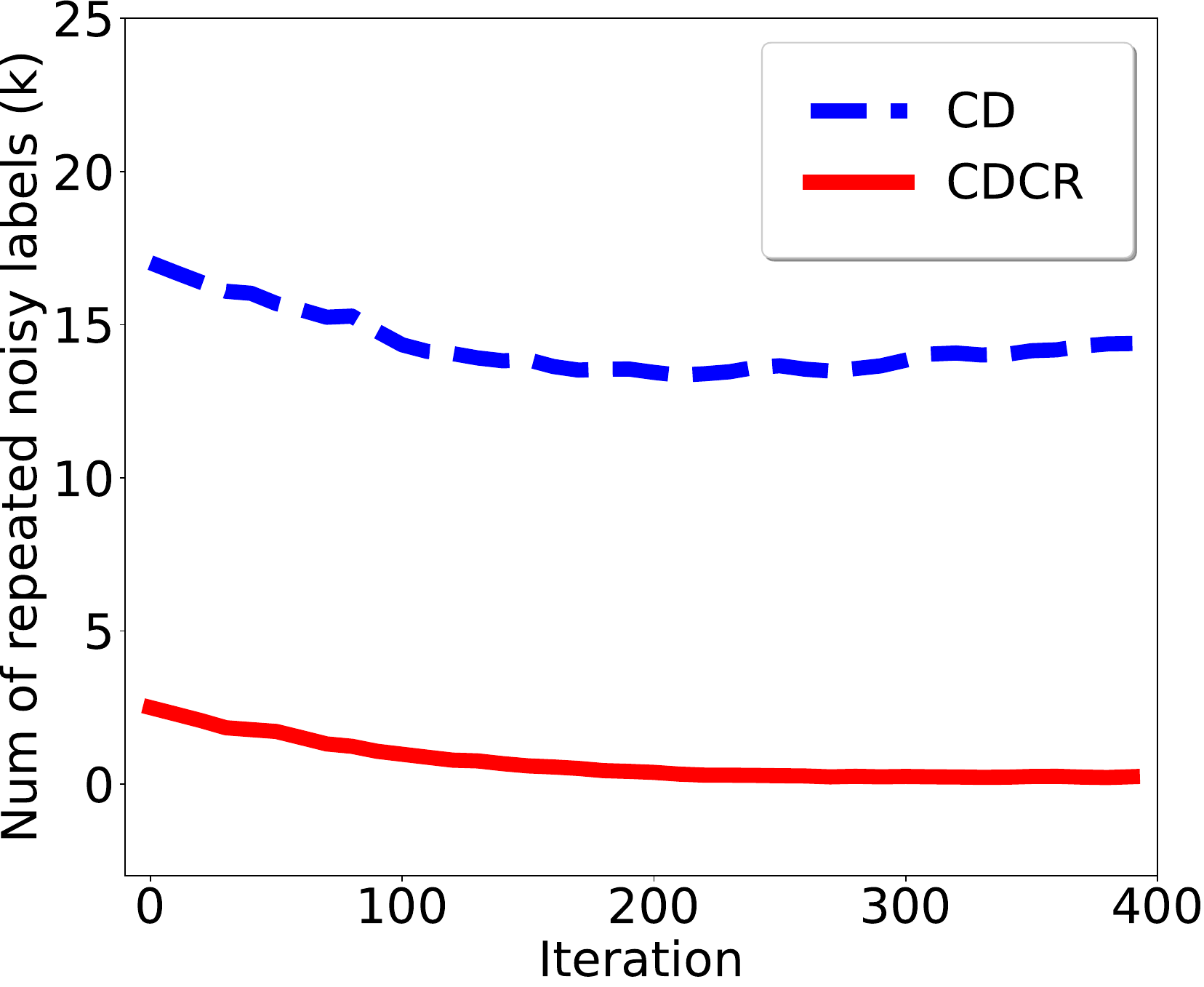}
	}
	
	\caption{Performance of different disambiguation strategies on VOC with flipping rate $q=0.4$. Experimental results show that CDCR achieves much better disambiguation performance than CD due to its strong robustness based on consistency regularization against noisy labels in the identified labels.}
	\label{fig:voc_analysis}
\end{figure}

To mitigate this issue, motivated by the advantage function in reinforcement learning \cite{arulkumaran2017deep}, we propose a multi-label version to perform disambiguation by incorporating the varied difficulties of different classes. Let $b_j$ be the baseline probability for $j$-th class, which can be computed by averaging predicted probabilities for the possible positive labels with respect to $j$-th class. Specifically, $b_j$ can be computed as follows:
\begin{equation*}
	b_j= \sum_{i=1}^n\mathbb{I}(p(y_j|\x_i)>0.5, \tilde{y}_{ij}=1)p(y_j|\x_i),
\end{equation*}
where the condition that $p(y_j|\x_i)>0.5$ means that $y_j$ is likely to be a relevant label for $\x_i$, and $\tilde{y}_{ij}=1$ is used to make sure that $y_j$ is a candidate label for $\x_i$. Obviously, the smaller the baseline probability, the more difficult the semantic class. Then, the curriculum defined in Eq.\eqref{eq:curriculum} can be re-formulated as follows:
\begin{equation*}
	\omega^*_{ij} = {\mathbb I}(p_{ij}-b_j\ge\alpha-\bar{b}), \forall i\in[n], j\in\{j'|\tilde{y}_{ij'}=1\},
\end{equation*}
where $\bar{b}$ represents the baseline probability for all possible positive labels, that can be computed by averaging predicted probabilities for possible positive labels with respect to all classes.

\subsection{Consistency Regularization}
\label{sec:consistency}

Although the idea of CD strategy seems to be effective for identifying true labels in the candidate set, we observe that the proposed method does not always work in our experiments, especially in the case with a large size of the candidate set. Figure \ref{fig:precision} and Figure \ref{fig:repeat} respectively illustrate the precision of CD strategy for identified labels, i.e., the percentage of the relevant labels in the identified ones, and the number of repeated noisy labels between any two sequential iterations in the identified ones as the number of iterations increases in the VOC dataset. From the figure, it can be observed that the CD strategy obtains an unfavorable performance for identifying true labels in the candidate set (the precision is around 0.3 in most cases). This is mainly because the deep neural net can be easily over-fitting to the identified label, including a large part of noisy labels, especially in the early training stages, which make the CD strategy fail to precisely identify the true labels. The above discussion can be further validated by Figure \ref{fig:repeat}, where the number of repeated noisy labels in the identified ones is relative large (more than 15 $k$), which indicates that CD strategy suffers the issue of over-fitting to the noisy labels mixed in the identified ones. 
To mitigate this issue, inspired of recent studies in semi-supervised learning \cite{berthelot2019mixmatch,sohn2020fixmatch}, we perform disambiguation for candidate labels by introducing the consistency regularization between original images and their augmented versions to prevent the model from over-fitting to the noisy labels in the candidate set. Specifically, for each image $\x$, we denote $\text{Aug}(\x)$ as its augmented version, where $\text{Aug}(\cdot)$ is a stochastic augmentation that does not alter the semantic of the image, and denote $\p(\y|\text{Aug}(\x))$ as the corresponding predicted probability distribution. 
Then, the loss function defined in Eq.\eqref{eq:DC} can be re-written as follows:

\begin{equation}\label{eq:DC_consistency}
	L_{\text{CDCR}} = \frac{1}{n}\sum_{i=1}^n\sum_{j=1}^K\omega_{ij}\ell(\tilde{y}_{ij},p(y_j|\text{Aug}(\x_i))) +  \Gamma(\Omega;\lambda).
\end{equation}

Figure \ref{fig:voc_analysis} further illustrates the performance curves of CD strategy with consistency regularization (CDCR for short) for identifying true labels in the candidate set. From the figures, it can be observed that CDCR can achieve disambiguation for candidate labels with a high precision (higher than 0.9). This indicates that the CDCR strategy has a strong robustness against the over-fitting to the noisy labels mixed in the identified ones, which can be further validated by \ref{fig:repeat}, where the number of repeated noisy labels is much less than CD strategy without consistency regularization. Therefore, CDCR obtains a more powerful disambiguation ability to exploit much more potential relevant labels and thus achieves a better generalization performance. The main procedures of CDCR are summarized in Algorithm 
\ref{al:CDCR}.
\begin{algorithm}[!tb]
	\caption{Curriculum based Disambiguation with Consistency Regularization (CDCR)}
	\label{al:CDCR}
	\begin{algorithmic}[1]
		\STATE \textbf{Input:} the training image set $D$, the max epoch $E$, the max iteration $T$ and the batch-size $B$
		\STATE \textbf{Process:}
		\begin{ALC@g}
			\STATE Warm-up the model with a few epochs.
			\STATE \textbf{For} $e=1, 2, \dots, E$ \textbf{do}
			\begin{ALC@g}
				\STATE Estimate the label weights $\Omega$ with Eq.\eqref{eq:curriculum}.
				\STATE \textbf{For} $t=1, 2, \dots, T$ \textbf{do}
				\begin{ALC@g}
					\STATE Sample a mini-batch $\{\x
				_i\}_{i=1}^{B}$.
				\STATE Perform augmentation for each image and obtain $\{\text{Aug}(\x_i)\}_{i=1}^{B}$.
					\STATE Update the model parameters $\btheta$ with Eq.\eqref{eq:DC_consistency}.
				\end{ALC@g}
				\STATE \textbf{End For}
				
			\end{ALC@g}
			\STATE \textbf{End For}
		\end{ALC@g}
		\STATE \textbf{Output:} the optimal model parameters $\btheta^*$  and label weights $\Omega^*$.
	\end{algorithmic}
\end{algorithm}

\section{Experiments}
\label{sec:experiment}

\subsection{Experimental Setting}

We perform experiments on two popular multi-label image classification benchmarks, i.e., MS-COCO \cite{lin2014microsoft} and VOC2007 \cite{everingham2010pascal}. MS-COCO contains a training set with 82,081 images and a validation set with 40,504 images, which can be classified into 80 different categories, with an average of 2.9 labels per image. PASCAL Visual Object Classes Challenge (VOC2007) is another widely used dataset for multi-label image classification. It is divided into a trainval set of 5,011 images and a test set of 4,952 images, which are from 20 object categories.
To construct candidate label sets for training images, each irrelevant label can be flipped into a candidate one with a probability $q$ sampled from $\{0.1, 0.2, 0.4\}$ for VOC2007 and $\{0.05, 0.1, 0.2\}$ for MS-COCO. 

Following the existing works \cite{chen2019multi,ridnik2021asymmetric}, we adopt the mean average precision (mAP) over all classes, average per-class precision, recall, F1-score (CP, CR, CF1) and overall precision, recall, F1-score (OP, OR, OF1) to evaluate the performance. Among these metrics, \textbf{mAP}, average overall F1-score (\textbf{OF1}), and average per-class F1-score (\textbf{CF1}) are more important, since they take both false-negative and false-positive rate into consideration. 

\begin{table*}[t]
	\centering
	\caption{Comparison results between the proposed method and comparing methods on the VOC2007 dataset. All metrics are in \%. The best performances are highlighted in bold.}   
	
	\label{table: voc comparison result}
	\begin{tabular}{l  | c c c | c c c | c c c}
		\toprule
		\multirow{2}*{Method} &
		\multicolumn{3}{c|}{$q=0.1$} &  \multicolumn{3}{c|}{$q=0.2$} & \multicolumn{3}{c}{$q=0.4$} \\
		\cmidrule(lr){2-10}
		& MAP & CF1 & OF1 & MAP & CF1 & OF1 & MAP & CF1 & OF1\\ 
		\bottomrule
		\midrule
		PMLNI & 80.32 &  65.64 &  68.56 & 77.63 &  63.44 &  67.26 & 69.95 &  58.21 &  63.39\\
		fPML  & 70.36 &  61.49 &  66.01 & 75.29 &  64.41 &  66.83 & 63.22 &  55.20 &  60.70\\
		PMLLRS & 80.68 &  66.10 &  68.72 & 77.08 &  62.99 &  66.96 & 63.86 &  55.45 &  61.01\\
		PARMAP &  69.61 &  66.69 &  67.12 & 67.66 &  64.43 &  66.38 & 65.69 &  61.18 &  64.45 \\
		PARVLS & 72.50 &  64.64 &  65.40 & 70.54 &  63.56 &  65.12 & 69.25 &  62.05 &  64.28\\
		\midrule
		BCE  & 84.04 &  68.46 &  70.09 & 80.43 &  63.86 &  68.63 & 71.35 &  59.95 &  65.05\\
		ASL & 86.62 &  68.50 &  72.29 & 83.89 &  67.34 &  71.07 &   74.99 &  62.54 &  67.52\\
		Query2Label & 87.14 &  68.10 &  72.82 & 84.81 &  66.75 &  71.72 & 79.90 &  63.88 &  68.84\\
		\midrule
		CDCR  & \textbf{90.12} &  \textbf{71.07} &  \textbf{72.96} & \textbf{88.94} &  \textbf{70.99} &  \textbf{73.00} & \textbf{86.15} &  \textbf{70.07} &  \textbf{71.94} \\
		\midrule
		\rowcolor{mygray}
		Supervised & 90.40 & 72.41 & 73.32 &- &- &- &- &- & -\\
		\bottomrule
	\end{tabular}
\end{table*}

To evaluate the effectiveness of the proposed method, we compare our method CDCR with five state-of-the-art PML methods: PML-NI \cite{xie2021partial}, PML-LRS \cite{sun2019partial}, fPML \cite{yu2018fpml}, PARMAP \cite{zhang2020pml}, PARVLS \cite{zhang2020pml} and two state-of-the-art multi-label image classification methods, ASL~\cite{ridnik2021asymmetric} and Query2Label~\cite{liu2021query2label}. Besides, we also compare with the baseline method (denoted by BCE) that directly trains a model on partial labels with the standard multi-label BCE loss.

Following works~\cite{ridnik2021asymmetric,liu2021query2label}, we employ TResNet-L~\cite{ridnik2021tresnet} pre-trained on ImageNet~\cite{deng2009imagenet} as our classification model, since it performs better than ResNet101~\cite{he2016deep} for multi-label image classification tasks while keeping similar efficiency constraints on GPU. We resize each image to $224\times224$ as the input resolution and adopt RandAugment~\cite{cubuk2020randaugment} and Cutout for augmentation. We adopt Adam optimizer and 1-cycle policy~\cite{smith2019super} to train the model for 80 epochs with maximal learning rate 0.0002. For the curriculum parameter $\alpha$, we set $\alpha=0.6$ for MS-COCO and $\alpha=0.8$ for VOC2007. In Section \ref{sec:sensitivity}, we further conduct experiments to analyze parameter sensitivity. 
The batch size is set as 64 for VOC2007 and 128 for MS-COCO. 
To make a fair comparison, we first adopt the TResNet-L pre-trained on ImageNet to extract features for the traditional PML methods. 
Moreover, we apply exponential moving average (EMA) to model parameters with a decay of 0.9997.  

\begin{table*}[t]
	\centering
	\caption{Comparison results between the proposed method and comparing methods on the MS-COCO dataset. All metrics are in \%. The best performances are highlighted in bold.}  
	\setlength{\tabcolsep}{ 1mm} 
	\label{table: COCO comparison result}
	\begin{tabular}{l  | c c c | c c c |c c c}
		\toprule
		\multirow{2}*{Method} &
		\multicolumn{3}{c|}{$q=0.05$} &  \multicolumn{3}{c|}{$q=0.1$} &\multicolumn{3}{c}{$q=0.2$}\\
		\cmidrule(lr){2-10}
		& MAP & CF1 & OF1 & MAP & CF1 & OF1 & MAP & CF1 & OF1\\ 
		\bottomrule
		\midrule
		PMLNI & 60.36 &  52.97 &  58.70 & 59.87 &  52.61 &  58.56 & 58.92 &  51.81 &  58.20 \\
		fPML  & 57.93 &  51.29 &  57.55 & 57.75 &  51.08 &  57.54 & 57.14 &  50.86 &  57.47 \\
		\midrule
		BCE  & 72.39  & 68.17  & 72.47 & 70.43 &  64.41 &  69.30 & 67.30 &  48.80 &  56.02 \\
		ASL & 76.92  & 61.11  & 66.98 & 75.14 &  60.00 &  66.26 & 72.28 &  57.65 &  64.88 \\
		Query2Label & 76.56 &  60.83 &  66.69 &  75.39 &  60.50 &  66.30 &73.59 &  59.36 &  65.54 \\
		\midrule
		CDCR  & 77.95 &  73.48 &  77.42 &  77.14 &  72.99 &  77.00 &  76.13 &  72.58 &  76.58  \\
		CDCR w/Diff &
		\textbf{78.05} &  \textbf{73.74} & \textbf{77.62} &  \textbf{77.35} &  \textbf{73.33} &  \textbf{77.22} &  \textbf{76.39} &  \textbf{72.80} & \textbf{76.79} \\
		\midrule
		\rowcolor{mygray}
		Supervised & 80.23 & 75.03 & 77.75 & - & - & - & - & - & - \\
		\bottomrule
	\end{tabular}
\end{table*}

\subsection{Results on VOC}

Table \ref{table: voc comparison result} reports comparison results between the proposed CDCR and the comparing methods on VOC2007 with flipping rate $q$ varying in $\{0.1,0.2,0.4\}$ in terms of mAP, CF1, and OF1. (Complete comparison results for all metrics are provided in the appendix). From the table, we can see that: 1) Our method significantly outperforms all the comparing methods by a significant margin in all cases. Specifically, we improve upon the state-of-the-art performance by $\bf{2.98}\%$,  $\bf{4.12}\%$ and  $\bf{6.25}\%$ in term of mAP. 2) Our method consistently achieves superior results as the size of the candidate set increases, while the comparing methods suffer a significant performance drop. 3) Our method achieves comparable results to the fully supervised learning model, in the case with $q=0.1, 0.2, 0.3$, where the differences between CDCR method and the fully-supervised model are respectively $\bf{0.38}\%$, $\bf{1.46}\%$, and $\bf{2.77}\%$ in terms of mAP, which validates that CDCR method sufficiently achieves disambiguation for candidate labels. 4) The traditional PML methods achieve unfavorable performances, even worse than the baseline BCE method, since these methods cannot train deep models by fine-tuning the powerful pre-trained models.

\subsection{Results on MS-COCO}
Table \ref{table: COCO comparison result} reports comparison results between the proposed CDCR and comparing methods with flipping rate $q$ varying in $\{0.05,0.1,0.2\}$. We do not report performances of traditional PML methods PMLLRS, PARMAP, and PARVLS, since they achieve undesirable performances in VOC2007 and require unbearable training time (more than 48 hours). From the results, it can be observed that: 1) Our CDCR method achieves the best performance in all cases. 2) By incorporating the difficulties of different classes, CDCR w/ Diff obtains significant improvements, especially in the cases with large flipping rates $q=0.1,0.2$, since CDCR w/ Diff benefits from identifying labels for difficult classes, which leads model to obtain a better generalization performances. 3) Our method consistently achieves superior results as the number of candidate labels increases, while the comparing methods suffer a significant performance drop. These results convincingly examine the effectiveness of the proposed method for solving PMLC tasks.

\subsection{Ablation Study}

In this section, to further analyze how the proposed method improves performance for PMLC tasks, we conduct a set of ablation studies on VOC2007 with flipping rate $q=0.4$ and report the results in Table \ref{table: ablation study on VOC2007}. We first conduct a simple experiment to show the importance of fine-tuning without any disambiguation technique. The achieved mAP is 71.35\% by fine-tuning on pre-trained models that is significantly better than 18.85\% achieved without fine-tuning. This may be one of main reasons for the traditional disambiguation methods that fail to obtain a desirable performance. To validate the effectiveness of the proposed curriculum based disambiguation strategy, we compare with the method (with fine-tuning and consistency regularization) that simply performs disambiguation once after the warm-up procedure. It can be observed that by performing disambiguation with curriculum, the mAP performance is improved to 85.71\% with 8.09\% increment. Finally, we also examine the usefulness of the consistency regularization. By using the consistency regularization, the mAP performance is improved with 16.13\% increment from 69.58\%. The main reason is that the consistency regularization provides the model with strong robustness against noise labels mixed in the identified ones. These results verify
that each component in the proposed method contributes to the performance improvement.

\begin{table*}[t]
	\centering
	\caption{Ablation study on VOC2007 with flipping rate $q=0.4$.}   
	\setlength{\tabcolsep}{ 1mm}
	\label{table: ablation study on VOC2007}
	\begin{tabular}{c c c  | c c c c c c c}
		\toprule
		Fine-tuning &  Consistency  & Curriculum & mAP & CP & CR & CF1 & OP & OR & OF1\\
		\bottomrule
		\midrule
		&  &  &  18.85 & 20.89 & 22.91 & 21.85 &  32.48 & 42.57 & 36.85\\
		$\checkmark$&  &  &  71.35 & 51.29 & 72.13 & 59.95 &  57.36 & 75.12 & 65.05\\
		$\checkmark$&  $\checkmark$&  &  77.62 & 53.12 & 75.63 & 62.41 &  59.95 & 78.07 & 67.82\\
		$\checkmark$&  &  $\checkmark$&  69.58 & 48.83 & 69.82 & 57.47 &  56.20 & 73.20 & 63.59\\
		\midrule
		$\checkmark$&  $\checkmark$&  $\checkmark$& 86.15  &61.87 & 80.77 & 70.07  &63.58 & 82.82 & 71.94\\
		\bottomrule
	\end{tabular}
\end{table*}

\subsection{Step-Wise Evaluation of Our Approach}
\label{sec:step-wise}

\begin{figure}[!tb]
	\centering
	
	\subfigure[CD vs BCE]{
		\label{fig:1_CD_BCE}
		\centering
		\includegraphics[width=0.9\linewidth]{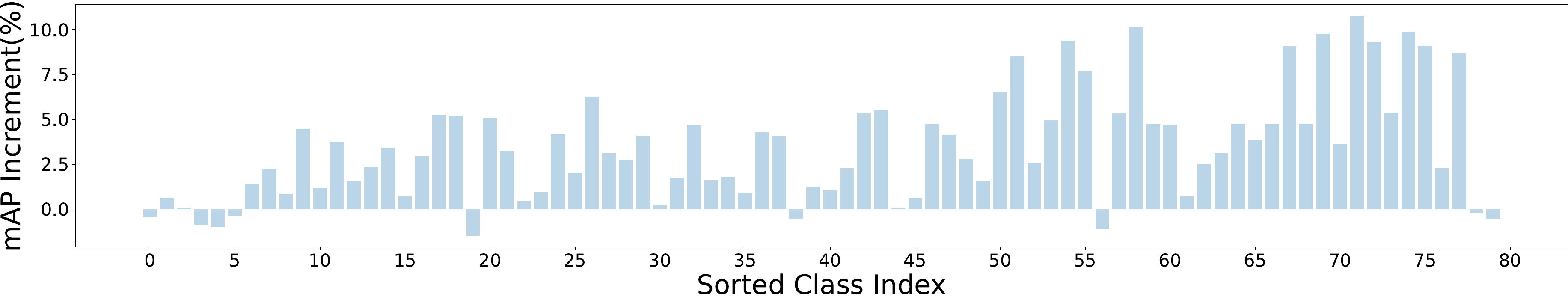}
	} 
	
	\subfigure[CDCR vs CD]{
		\label{fig:2_CDCR_CD}
		\centering
		\includegraphics[width=0.9\linewidth]{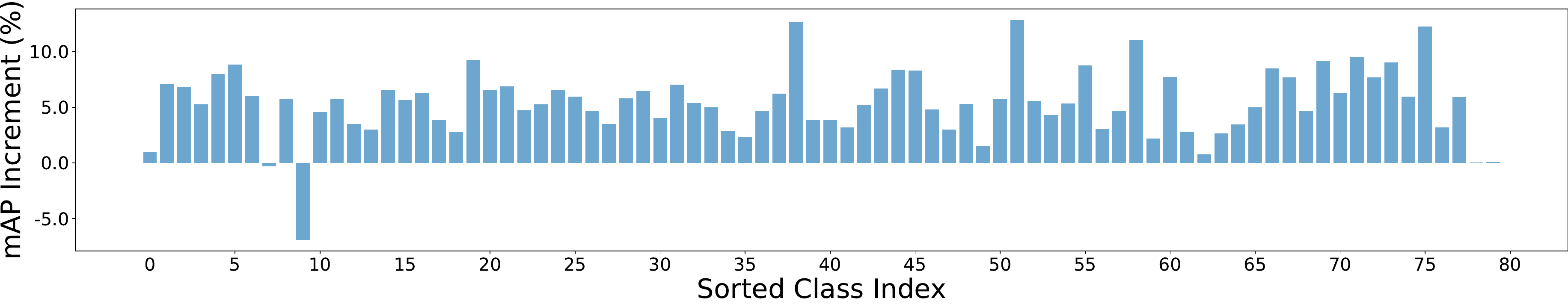}
	}
	
	\subfigure[CDCR w/ Diff vs CDCR]{
		\label{fig:3_CDCRCD_CDCR}
		\centering
		\includegraphics[width=0.9\linewidth]{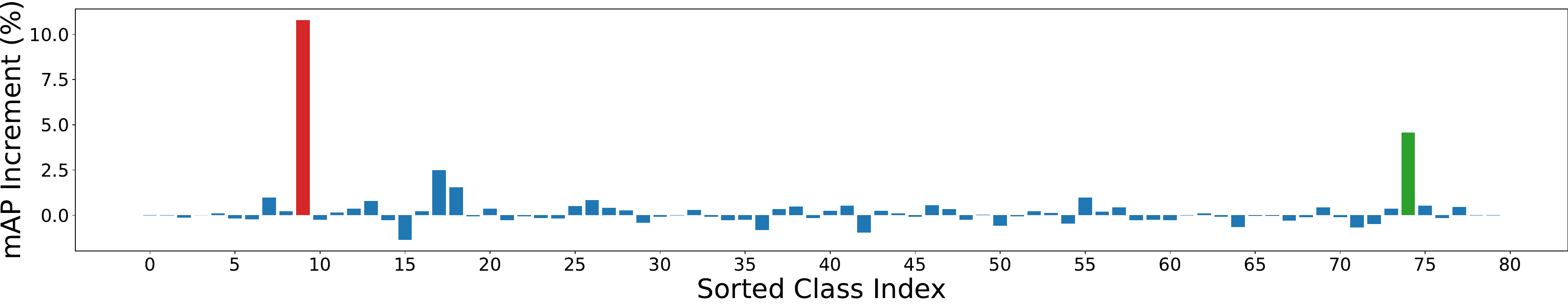}
	}
	
	\caption{Corresponding to one baseline and three strategies, i.e. BCE, curriculum based disambiguation, consistency regularization and class difficulty, we show the per-class mAP increments between each two strategies to evaluate our method. The class index is sorted from left to right based on the noise rate.}
	\label{fig:step-wise}
\end{figure}

We perform a step-wise evaluation on the COCO with flipping rate $q=0.2$ by showing mAP increment per-class to have a better understanding of how curriculum disambiguation strategy, consistency regularization and class difficulty work on different classes sorted in ascending order according to the noise rate, i.e., the percentage of false positive labels hidden in observed positive labels for each class. It is noteworthy that since the flipping rate $q$ is identical for all classes, for each class, a smaller number of relevant labels leads to a larger number of false positive labels, which indicates a larger value of noise rate.
Figure \ref{fig:1_CD_BCE} shows the proposed curriculum based disambiguation achieves a significant improvement on almost all classes when compared to BCE. Figure \ref{fig:2_CDCR_CD} illustrates the performance of the consistency regularization on the basis of curriculum based strategy. From the figure, it can be observed that the consistency regularization can further boost the model performance, which demonstrates that its powerful ability to prevent the model from over-fitting to the noisy labels in the candidate set. Finally, figure \ref{fig:3_CDCRCD_CDCR} reports the improvements of performance by further incorporating the class difficulty. In the figure, the class
colored by red corresponds to a hard class (since CDCR achieves a relative low AP with respect to this class. See the detailed results in appendix.) while the class  
colored by green corresponds to a tail class with few positive examples. In particular, these results show that the disambiguation performances of either hard classes or tail classes can be improved by further considering the class difficulty, which provides an empirical validation for the discussion in Section \ref{sec:cd}.

\subsection{Parameter Sensitivity Analysis}
\label{sec:sensitivity}
\begin{figure}[!tb]
	\centering
	
	\subfigure[VOC2007]{
		\label{fig:alpha_voc}
		\centering
		\includegraphics[width=0.45\linewidth]{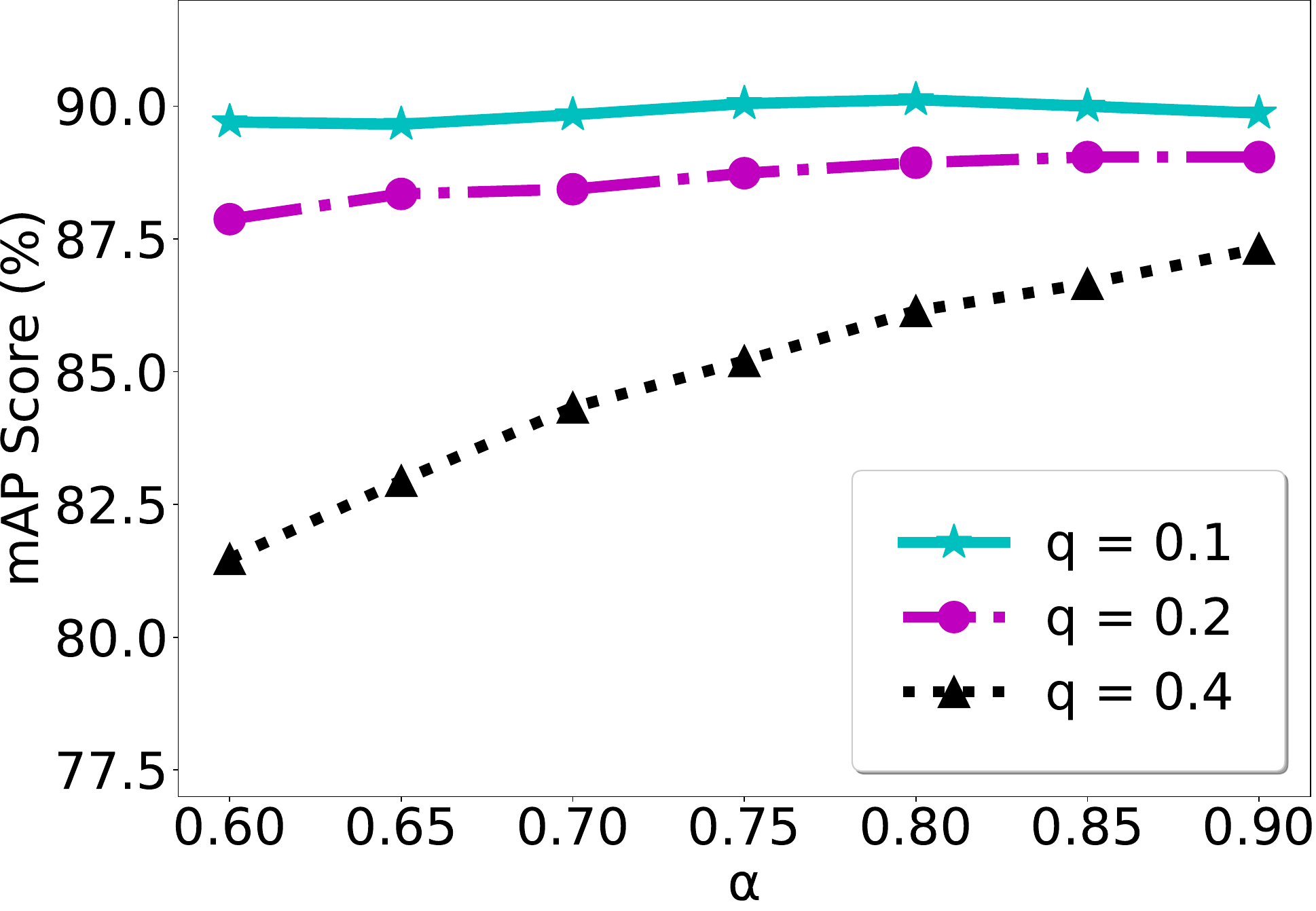}
	} 
	\subfigure[COCO]{
		\label{fig:alpha_coco}
		\centering
		\includegraphics[width=0.45\linewidth]{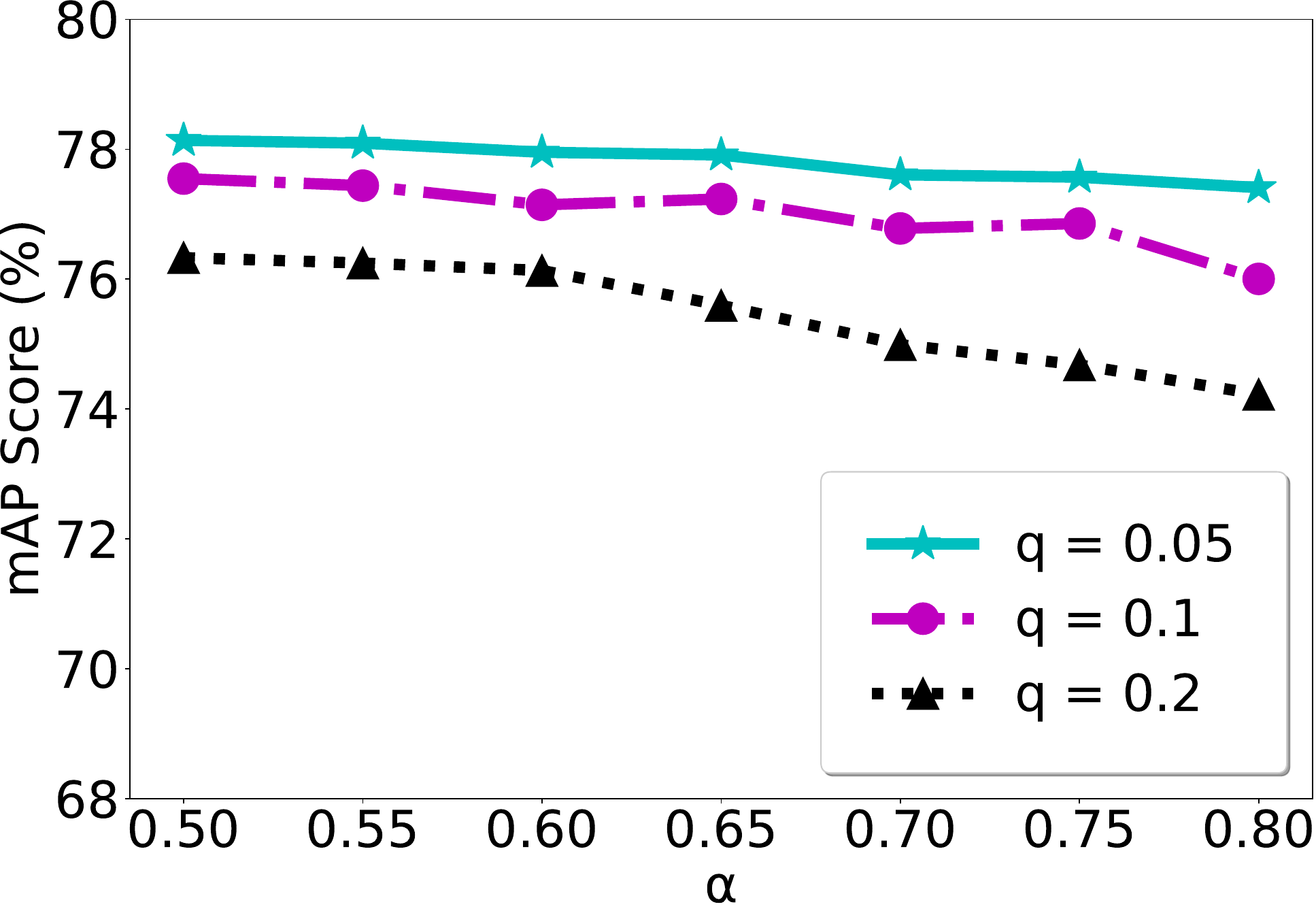}
	}
	
	\caption{Accuracy comparisons with different values of $\alpha$ on VOC2007 and COCO}
	\label{fig:alpha}
\end{figure}

Finally, we study the influence of the curriculum parameter $\alpha$ on the performance of CDCR. Figure \ref{fig:alpha_voc} illustrates the mAP curves of CDCR as the value of $\alpha$ changes among $\{0.60, 0.65, 0.70, 0.75, 0.80, 0.85, 0.90\}$ on VOC2007 with flipping rate $q$ varying in $\{0.1, 0.2, 0.4\}$. From the figure, it can be observed that the performance of CDCR is not very sensitive to the parameter $\alpha$, especially in the case with small flipping rates, i.e., $q=0.1, 0.2$. Regarding the cases with large flipping rates $q=0.4$, CDCR can still obtain the desirable performances if the value of $\alpha$ is large enough, since a large value of threshold $\alpha$ is beneficial for identifying true labels as mentioned in Section \ref{sec:cd}. Figure \ref{fig:alpha_coco} illustrates the mAP curves of CDCR as the value of $\alpha$ changes among $\{0.50, 0.55, 0.60, 0.65, 0.70, 0.75, 0.80\}$ on COCO with flipping rate $q$ varying in $\{0.05, 0.1, 0.2\}$. From the figure, the performance of CDCR  decreases as the value of $\alpha$ becomes larger, which is mainly due to the fact that the average number of relevant labels for each image is much larger in COCO dataset. Therefore, as the threshold $\alpha$ increases, the number of identified true labels gradually decreases and thus degrades the disambiguation performance, which causes negative impact on the model performance.

\section{Conclusion}
\label{sec:conclusion}

In this paper, we make a first attempt to learn a deep neural network for solving PMLC problems, where each image is assigned with a candidate label set consisting of multiple relevant labels and other noisy labels. In the proposed method, curriculum based disambiguation strategy is designed to 
progressively identify true labels in the candidate set by considering varied difficulties of different classes. Furthermore, the consistency regularization is utilized to retrain the model by using stochastic augmentations for training images, which leads the model to obtain a strong disambiguation ability. Extensive experiments on the commonly used benchmark datasets show the proposed method can achieve state-of-the-art performances.

\bibliographystyle{unsrt}  
\bibliography{egbib}

\end{document}